%% file: interspeech_main.tex
\documentclass[a4paper]{article}

\usepackage{amsmath}
\usepackage{graphicx}
\usepackage[utf8]{inputenc} 
\usepackage[T1]{fontenc}    
\usepackage{hyperref}       
\usepackage{url}            
\usepackage{booktabs}       
\usepackage{amsfonts}       
\usepackage{nicefrac}       
\usepackage{microtype}      
\usepackage{lipsum}
\usepackage{graphicx}
\usepackage{colortbl}

\usepackage{multicol}
\usepackage{multirow}
\usepackage{booktabs}
\usepackage[table]{xcolor}
\usepackage[inline]{enumitem}
\usepackage{subcaption}
\usepackage{tabularx}
\usepackage{multirow, makecell}
\usepackage{siunitx}
\usepackage[nocompress]{cite}

\input{macro}

\usepackage{INTERSPEECH2022}

\title{Wav2Vec-Aug: Improved self-supervised training with limited data}

\name{Anuroop Sriram, Michael Auli, Alexei Baevski}
\address{Facebook AI}

\email{\{anuroops,michaelauli,abaevski\}@fb.com}

\begin{document}

\maketitle
 
\begin{abstract}
Self-supervised learning (SSL) of speech representations has received much attention over the last few years but most work has focused on languages and domains with an abundance of unlabeled data.
However, for many languages there is a shortage even in the unlabeled data which limits the effectiveness of SSL.
In this work, we focus on the problem of applying SSL to domains with limited available data by leveraging data augmentation for Wav2Vec 2.0 pretraining.
Further, we propose improvements to each component of the model which result in a combined relative word error rate (WER) improvement of up to 13\% compared to Wav2Vec 2.0 on Librispeech test-clean / other.
\end{abstract}
\noindent\textbf{Index Terms}: self-supervised learning, data augmentation, contrastive learning

\input{sections/intro}
\input{sections/bg}

\input{sections/dataaug}

\input{sections/arch}

\input{sections/expts}

\input{sections/results}

\input{sections/conclusion}

\bibliographystyle{IEEEtran}

\bibliography{mybib}

\end{document}

%% file: macro.tex
\newcommand{\Lab}{LS-Lab}

\newcommand{\Inp}{\mathcal{X}}
\newcommand{\Feat}{\mathcal{Z}}
\newcommand{\QFeat}{\mathcal{Q}}
\newcommand{\Context}{\mathcal{C}}

\DeclareMathOperator{\CMLP}{CMLP}
\DeclareMathOperator{\TMLP}{TMLP}
\DeclareMathOperator{\U}{Uniform}

\newcommand{\xe}{\mathbf{x}}
\newcommand{\ze}{\mathbf{z}}
\newcommand{\zq}{\mathbf{q}}

\newcommand{\cc}{\mathbf{c}}

%% file: sections/intro.tex
\section{Introduction}
\label{sec:intro}

The use of self-supervised learning (SSL) for learning representations from unlabeled speech has received much attention over the last few years~\cite{oord2018cpc,schneider2019wav2vec,harwath2019learning,chung2019apc,pascual2019learning}. 
However, much of this research assumes that a large amount of unlabeled audio data is available, which is not always the case for rare languages or for specialized domains. 
The performance of SSL for speech with limited unlabeled data has been less studied.

Data augmentation has proven to be an effective strategy for supervised learning~\cite{ko2015audio,amodei2016deepspeech} when the amount of labeled data is limited. Kharitonov et al.\cite{kharitonov2021dataug} showed that data augmentation benefits Contrastive Predictive Coding (CPC; \cite{oord2018cpc}) when a limited amount of unlabeled data is available.
CPC is light-weight self-supervised learning algorithm which uses past context to make predictions about the future, and we show that data augmentation also works well for stronger transformer based bi-directional models such as Wav2Vec 2.0~\cite{baevski2020wav,conneau2020unsupervised}.

We also propose improvements to each component of the Wav2Vec 2.0 model that are useful for both the small data setting as well as the large data settings. 
First, we replace some of the convolutional layers of the feature encoder in Wav2Vec 2.0 with improved layers based on light and dynamic convolutions \cite{wu2019pay}. 
Second, we replace the transformer part of the model with conformer layers, similar to~\cite{gulati2020conformer,zhang2020pushing}. 
Finally, we add MLP projections on top of the context and latent vectors of Wav2Vec 2.0 inspired by self-supervised learning algorithms in the computer vision literature~\cite{he2019momentum,chen2020simple}.
Together, these changes improve the performance on downstream ASR task by up to 13\% (relative) WER compared to a Wav2Vec 2.0 baseline. We call this model \emph{Wav2Vec-Aug}.


%% file: sections/bg.tex
\section{Background}
\label{sec:system}

\begin{figure}[htb]
  \centerline{\includegraphics[width=8cm]{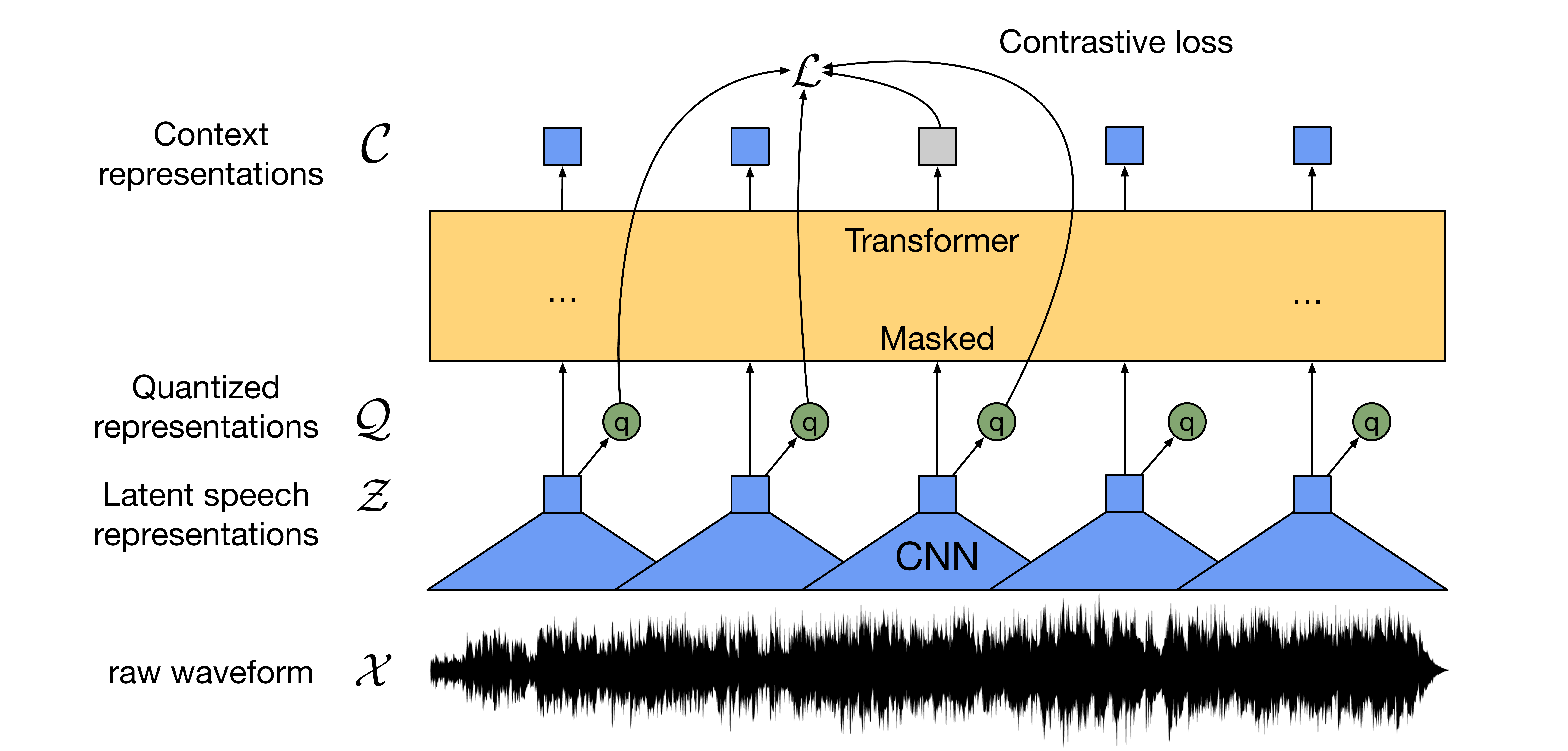}}
\caption{The Wav2Vec 2.0 model}
\label{fig:w2v}
\end{figure}

Our model is based on Wav2Vec 2.0 (W2V2;~\cite{baevski2020wav,conneau2020unsupervised}) which maps raw audio~$\xe \in \Inp$ to a latent feature representation $\cc_1, \ldots, \cc_T$. The Wav2Vec 2.0 model consists of a convolutional feature encoder $f: \Inp \mapsto \Feat$ that first maps the input audio to a latent speech representation $\ze_1, \ldots \ze_T$. This latent representation is then input to a Transformer model $g: \Feat \mapsto \Context$ to output the context representations $\cc_1, \ldots, \cc_T$~\cite{baevski2019vqwav2vec,baevski2019effectiveness}. Each $\ze_t$ represents about 25ms of audio strided by 20ms and the Transformer architecture follows the BERT model~\cite{vaswani2017transformer,devlin2018bert}. The model architecture is shown in Figure \ref{fig:w2v}.

In the pre-training phase, latent representations are discretized to $\zq_1, \dots, \zq_T$ with a quantization module $\Feat \mapsto \QFeat$ to represent the targets in the objective. The quantization module uses a Gumbel softmax~\cite{jang2017categorical} to choose entries from $G=2$ codebooks with $V=320$ entries each and the chosen entries are concatenated to obtain $\zq$~\cite{jegou2011ieee,jang2016gumbel,baevski2019vqwav2vec}. The model is trained to identify the true quantized latent $\zq_t$ using $\cc_t$ for each masked time-step within a set of $K=100$ distractors $\mathbf{Q}_t$ sampled from other masked time steps.

%% file: sections/dataaug.tex
\section{Data Augmentation}
\label{sec:augmentation}


Prior research has shown that data augmentation on input audio can be helpful for supervised learning \cite{ko2015audio,amodei2016deepspeech}, particularly in the small data setting.
This has been confirmed for self-supervised learning using the CPC algorithm~\cite{kharitonov2021dataug}.
In this work, we experiment with three different data augmentation methods for self-supervised learning: additive augmentation, pitch shift and reverberation. We have also experimented with speed perturbation, but found that it does not improve accuracy.

Additive augmentation involves adding a noise signal $\xe'$ to the input signal $\xe$. The noise signal can be chosen uniformly at random from a large collection of audio signals. The chosen noise signal is added at an SNR value of $s \sim \U(s_0, s_1)$, for given hyperparameters $s_0$ and $s_1$.

Pitch shift involves raising or lowering the pitch of the input audio by some random factor $f$. In our model, we sample the pitch shift factor from a gaussian distribution: $f \sim \mathcal{N}(0, \sigma_p)$, where $\sigma_p$ is a hyperparameter. Reverberation involves simulating far-field speech by convolving the input audio signal with a randomly generated room impulse response (RIR). The room size parameter $r$  was chosen randomly by sampling a number $r' \sim \mathcal{N}(0, \sigma_r)$, and then setting $r = min(|r'|, 100).$

During pre-training, for each data sample, we choose whether or not to apply each augmentation method independently with a probability $p$. This ensures that every combination of augmentations is applied to some training example. 

We first duplicate the input audio $\xe$ into source audio $\xe^{(s)}$ and target audio $\xe^{(t)}$. We then use $\xe^{(s)}$ to generate the context vectors, and $\xe^{(t)}$ to generate the target latent representations $\zq_t$. In our experiments, we found that it is beneficial to apply different augmentations to the source and target audios.

We used the WavAugment toolkit\footnote{https://github.com/facebookresearch/WavAugment} \cite{kharitonov2021dataug} to implement all augmentation methods.

%% file: sections/arch.tex
\section{Architectural Improvements}
\label{sec:architecture}

\subsection{Lightweight and Dynamic Convolution}
\label{sec:lightconv}
The Wav2Vec 2.0 model uses a purely convolutional feature encoder to extract features from the input audio signal. Wu et al. \cite{wu2019pay} introduced lightweight and dynamic convolutional layers that have been shown to outperform standard convolutions for a range of tasks.\footnote{We refer to non-separable convolutions as standard convolutions.} 
Lightweight convolutions are depth-wise separable convolutions~\cite{howardMobileNets} that share weights across groups of $m$-channels ($m$ is treated as a hyper-parameter). Dynamic convolutions build on lightweight convolutions by dynamically computing the convolution kernels as a function of the input at a given timestep.

These layers are computationally more efficient than transformer layers, making them suitable for the feature encoder, which operates over many time steps. 
We replace the final $k$ convolutional layers of the feature encoder with lightweight or dynamic convolutions. 
We find that $k=2$ leads to improved performance for both light and dynamic convolutions (Section \ref{sec:experiments}).

\subsection{Conformer}
\label{sec:conformer}
The transformer part of the Wav2Vec 2.0 model uses a series of multi-head self-attention blocks \cite{vaswani2017transformer}. Recently, Conformer models~\cite{gulati2020conformer} have been shown to outperform the original transformer architecture for both supervised and self-supervised speech recognition\cite{gulati2020conformer,zhang2020pushing}.

The Conformer architecture consists of a series of multi-head self attentions, depth-wise convolution and feed-forward layers. 
This combination helps the Conformer make effective use of both global and local interactions which leads to improved performance. 

\subsection{Context and Target MLPs}
\label{sec:mlphead}

Our final improvement is to modify the computation of the loss values by adding multi-layer perceptrons to the network~\cite{he2019momentum,chen2020simple}.
Specifically, we introduce two multi-layer perceptrons (MLPs), dubbed \emph{ContextMLP} and \emph{TargetMLP}, which are applied to the context $\cc$ and latent vectors $\zq$, respectively:

\begin{align*}
    \cc' &= \CMLP(\cc) \\
    \zq' &= \TMLP(\zq) \\
\end{align*}

The original Wav2Vec 2.0 model was trained to identify the true $\zq_t$ from $\cc$ within a set of $K$ distractors $\Tilde{\zq}$ using a contrastive loss. The inputs to the contrastive loss are cosine similarities between the context vector $\cc$ with $\zq_t$ and $\Tilde{\zq}$. In our model, we apply the $\CMLP$ to $\cc$, and the $\TMLP$ to $\zq_t$ and $\Tilde{\zq}$ before computing the cosine similarities. This allows the network to learn more complex, non-linear similarity functions than a simple cosine similarity. Concurrent to our work, Wu et al. \cite{wu2021performanceefficiency} has also incorporated the use of Context MLPs in a Wav2Vec 2.0 style model.

%% file: sections/expts.tex
\section{Experimental Setup}

We implement our changes on top of the Wav2Vec 2.0 Base model architecture~\cite{baevski2020wav}, by modifying its various components.
We use the same training setup as the Wav2Vec 2.0 Base model, except that we reduce the effective batch size by training models on 48 GPUs instead of the 64 GPUS used in \cite{baevski2020wav}. We implemented our model in the fairseq library~\cite{ott2019fairseq} and use the Librispeech dataset~\cite{panayotov2015librispeech} for all of our experiments. Pre-training our model for 400K steps takes roughly 2.5 days with 64 V100 GPUs.

We test our model with varying amounts of pre-training and fine-tuning data by creating subsets of the Librispeech dataset. 
For pre-training, we consider a subset of 50 hours randomly sampled from the clean-100h set, the clean-100h set and the full set (960 hours). 
For fine-tuning, we use the 1 hour split of Libri-light~\cite{kahn2020librilight} (\Lab-1H) and clean-100h (\Lab-100H). 
We pre-train models for 150K, 200K or 400K steps for the 50 hour, 100 hour and full datasets, respectively and follow the same procedure for fine-tuning and beam search as~\cite{baevski2020wav}.

After pre-training, we fine-tune each model on the labeled datasets with Connectionist Temporal Classification~\cite{graves2006connectionist}. After fine-tuning, we decode the fine-tuned models using the word-level $4$-gram language model (LM) from \cite{panayotov2015librispeech} using the beam search decoder from~\cite{pratap2019wav2letter}. 
We use a Bayesian Optimization\footnote{\url{https://github.com/facebook/Ax}} procedure to find the best decoding hyper-parameters over 128 trials with the following search space: LM weight ($[0,8]$), word score ($[-5,5]$), and silence score ($[-5,5]$). We use a beam size of $500$ for the hyperparameter search. After finding the optimal hyperparameters, we increase the beam size to $1,500$ to obtain the final results.

\begin{table*}[t]
    \small
    \centering
    \caption{Comparison of our improvements to the original Wav2vec 2.0 model on the Librispeech benchmark for different amounts of training pre-training and fine-tuning data. The Wav2Vec-Aug model obtains 9-13\% improvement in WER compared to Wav2Vec 2.0 on the test datasets.}
        \begin{tabular}
            {c|c|l|*{4}S[table-format=4.3]}
            \toprule
            \multirow{2}{*}{Pre-train data} & \multirow{2}{*}{Fine-tune data} & \multirow{2}{*}{Model} & \multicolumn{4}{c}{WER} \\
            & & & {dev-clean} & {dev-other} & {test-clean} & {test-other} \\
            \midrule
            
            \multirow{2}{*}{50H} & \multirow{2}{*}{1H} & Wav2vec 2.0 & 15.73 & 30.23 & 16.14 & 31.96\\
                        &   & Wav2Vec-Aug & 13.37 & 25.57 & 14.04 & 28.03 \\
            \midrule
            \multirow{4}{*}{100H} & \multirow{2}{*}{1H} & Wav2vec 2.0 & 10.50 & 20.66 & 10.82 & 21.52\\
                        &    & Wav2Vec-Aug & 9.17 & 18.15 & 9.81 & 19.45 \\
                          \cmidrule{2-7}
                    & \multirow{2}{*}{100H} & Wav2vec 2.0 & 3.82 & 11.97 & 4.50 & 12.31 \\
                                &    & Wav2Vec-Aug & 3.65 & 10.34 & 4.34 & 11.37 \\
            \midrule
            \multirow{4}{*}{960H} & \multirow{2}{*}{1H} & Wav2vec 2.0 & 6.03 & 12.17 & 6.22 & 12.45\\
                        &    & Wav2Vec-Aug & 5.17 & 10.39 & 5.71 & 11.42 \\
                          \cmidrule{2-7}
                    & \multirow{2}{*}{100H} & Wav2vec 2.0 & 2.86 & 8.17 & 3.46& 8.27\\
                                &    & Wav2Vec-Aug & 2.53 &7.14 & 3.04 & 7.21 \\
            
            \bottomrule
        \end{tabular}
    \label{tab:final}
\end{table*}

%% file: sections/results.tex
\section{Results}
\label{sec:experiments}

\subsection{Performance of Wav2Vec-Aug}

We first compare the performance of Wav2Vec-Aug with all of the proposed modifications (data augmentation during pre-training, replacing the last two feature encoder layers with dynamic convolutions, context and target MLPs, and conformer blocks) with Wav2Vec 2.0 on different amounts of pre-training and fine-tuning data. 
Table \ref{tab:final} shows that our approach can obtain up to a 13\% relative WER improvement over the baseline on test-clean/other depending on the amount of available data.
In the rest of this section, we examine each component of our approach in more detail.

\begin{figure}[htb]

  \centerline{\includegraphics[width=8cm]{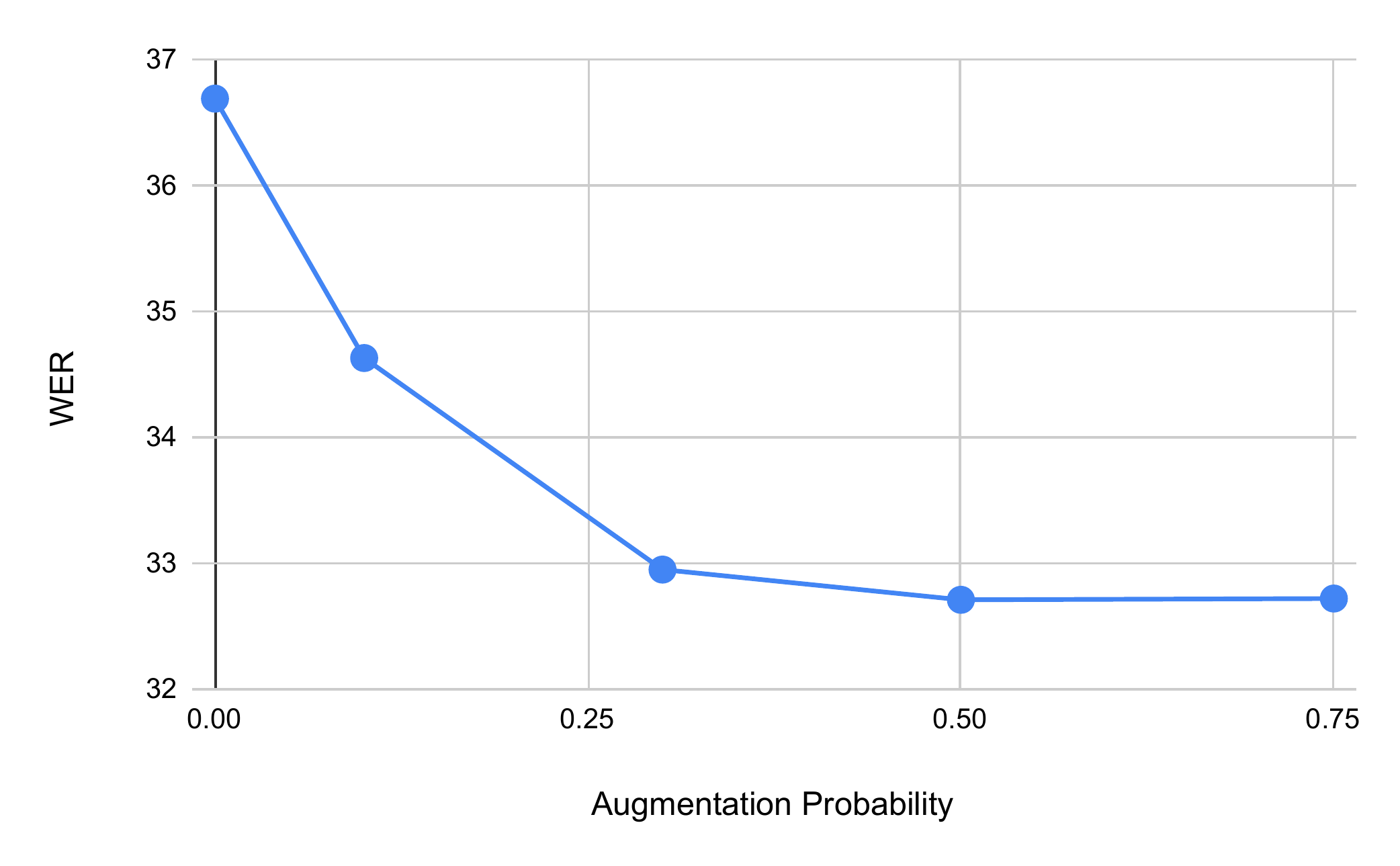}}
\caption{WER on dev-other for different augmentation probabilities. All models were pre-trained on the 50 hour subset with additive, pitch and reverb augmentations with $s_0 = 10, s_1 = 15, \sigma_p = 50,$ and $\sigma_r = 60$. Note that these experiments use a beam size of 5 and are not directly comparable to other results.}
\label{fig:data_aug_prob}

\end{figure}

\begin{figure}[htb]

  \centerline{\includegraphics[width=8.5cm]{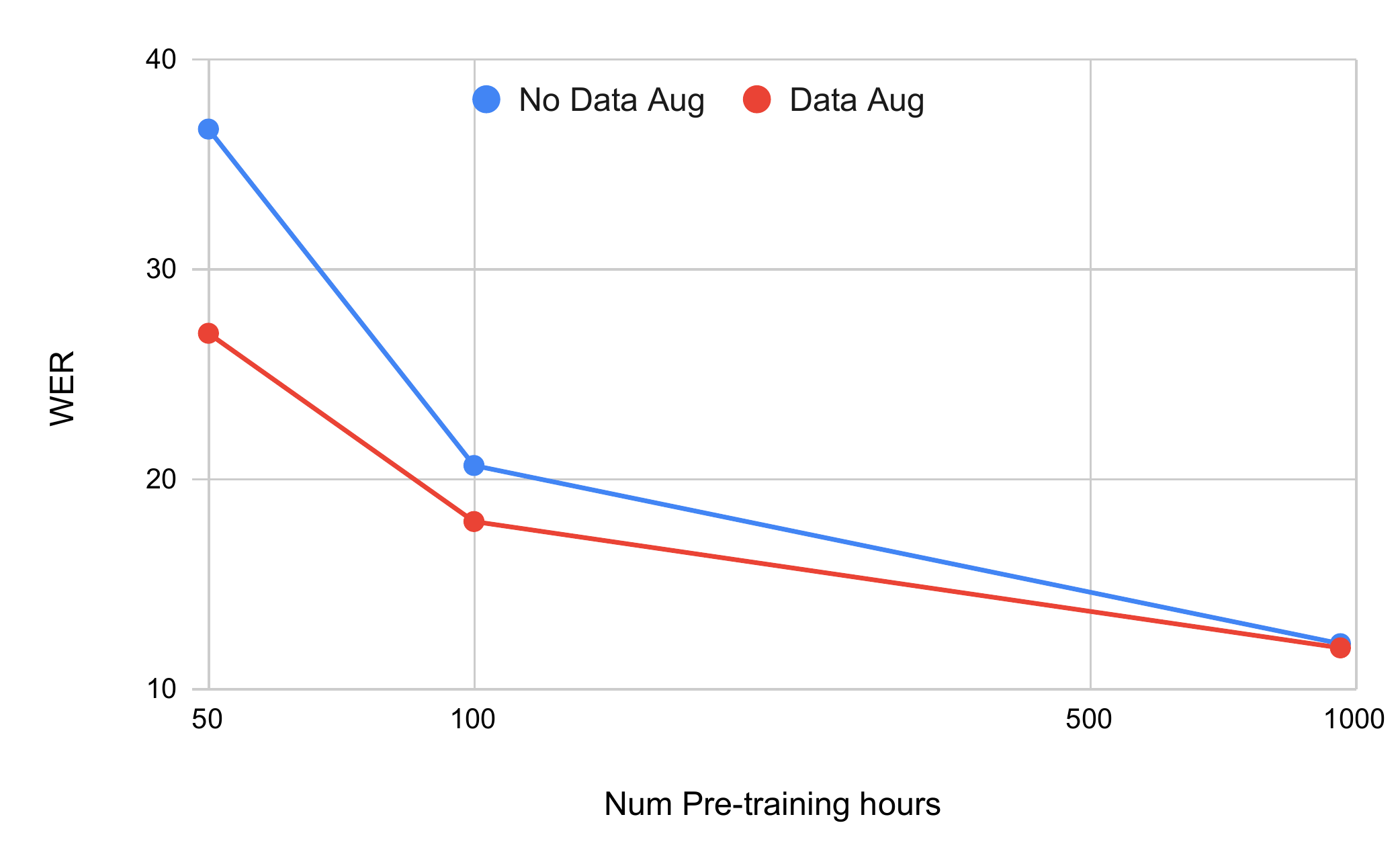}}
\caption{Effect of data augmentation during pre-training for different amounts of pre-training data. We show WER on the dev-other. 
The blue line shows the baseline model with no augmentation, while the red line shows results with the best augmentation probability $p = 0.5$. 
All models were fine-tuned on the 1 hour labeled subset of Librispeech.}
\label{fig:data_aug}
\end{figure}

\subsection{Data Augmentation}

We measure the effect of data augmentation vs. no data augmentation for different amounts of pre-training data when we use one hour of labeled data.
For additive augmentation, we used the MUSAN dataset~\cite{snyder2015musan} to draw the noise samples. 
Table \ref{tab:data_aug} shows results for different settings of the data augmentation hyperparameters. Based on these ablations, we set the augmentation parameters in our final model to $s_0 = 10$, $s_1 = 15$ for additive augmentation, $\sigma_p = 50$ as the standard deviation for pitch shift augmentation, and $\sigma_r = 60$ as the room size standard deviation for reverberation. Figure \ref{fig:data_aug_prob} shows the WERs for different augmentation probabilities when all three augmentation methods are used with these parameters. We obtain the best results with an augmentation probability of $p = 0.5$.

Figure~\ref{fig:data_aug} shows results of using data augmentation with these hyperparameters with different amounts of pre-training data using the same model architecture.
It is clear that data augmentation is most helpful in the low-data regime, improving WER by almost 11\% relative to the baseline model when only 50 hours of data is available for pre-training. These improvements begin to taper off as the size of the pre-training dataset is increased.

\begin{table}[t]
    \small
    \centering
    \caption{WER on dev-other for different data augmentation hyper-parameters for pre-training. All models were pre-trained on the 50 hour subset and then fine-tuned on the 1 hour labeled set. Note that these experiments use a beam size of 5 and are not directly comparable to other results.}
        \begin{tabular}
        {l|l|*{2}S[table-format=3.3]}
            \toprule
            Augmentation & Aug. Params & {Aug. Prob} & {WER} \\
            \midrule
            -- & -- & 0. & 36.69 \\
            \midrule
            Additive & $s_0=10, s_1=15$ & 0.5 & 33.37 \\
            \midrule
            \multirow{3}{*}{Pitch} & $\sigma_p = 20$ & 0.5 & 33.72 \\
              & $\sigma_p = 50$ & 0.5 & 33.56 \\
              & $\sigma_p = 100$ & 0.5 & 34.07 \\
            \midrule
            \multirow{3}{*}{Reverb} & $\sigma_r = 10$ & 0.5 & 38.30 \\
                                    & $\sigma_r = 25$ & 0.5 & 36.87 \\
                                    & $\sigma_r = 60$ & 0.5 & 36.02 \\
            \bottomrule
        \end{tabular}
    \label{tab:data_aug}
\end{table}

\subsection{Lightweight and Dynamic Convolution}

Table~\ref{tab:lconv} shows the results of replacing either the last two or the last four layers in the feature encoder with lightweight or dynamic convolutional layers. 
The Wav2Vec 2.0 model uses convolutional layers with $512$ channels in the feature encoder. To keep the number of parameters constant across experiments, we increased the number of channels to $640$ when the last two layers are changed, and to $608$ when the last four layers are changed. We used $8$ attention heads in the lightweight and dynamic convolution layers, and a dropout rate of $0.1$.

Replacing the last two layers yields a roughly 5\% relative improvement in WER on the dev-other validation set for both lightweight and dynamic convolutions, when fine-tuned on the 100 hour subset of Librispeech. 
Replacing more layers appears to be detrimental to the performance, which indicates that lightweight / dynamic convolutions are most useful in the final layers where the number of timesteps is relatively small.

\begin{table}[t]
    \small
    \centering
    \caption{WER on the dev-other dataset showing the effects of our proposed architectural improvements. All models were pre-trained on the full Librispeech dataset and fine-tuned on the 1 hour or 100 hour labeled sets. Each architectural improvement is applied independently to the Wav2vec 2.0-Base model.}
        \begin{tabular}{l|cc}
            \toprule
            \multirow{2}{*}{Modification} & \multicolumn{2}{c}{Dev-other WER} \\
            & \Lab-1H & \Lab-100H \\
            \midrule 
            Wav2vec 2.0 & 12.17 & 8.17 \\
            \midrule
            + Light Conv (last 2) & \textbf{11.73} & \textbf{7.80}\\
            + Light Conv (last 4) & 12.09 & 8.34\\
            \midrule
            + Dyn Conv (last 2) & \textbf{11.64} & \textbf{7.79}\\
            + Dyn Conv (last 4) & 12.17 & 8.23\\
            \midrule
            + Conformer &  \textbf{11.34} & \textbf{7.32} \\
            \midrule
            + $\CMLP$ (2 layers) & 11.43 & 8.04\\
            + $\TMLP$ (2 layers) & 11.50 & 8.12\\
            + $\CMLP +\TMLP$ \\ 
            \hspace{1em} 2 layers & 11.36 & 7.98\\
            \hspace{1em} 3 layers & 10.76 & 7.82\\
            \hspace{1em} 4 layers & \textbf{10.46} & \textbf{7.71}\\
            \bottomrule
        \end{tabular}
    \label{tab:lconv}
\end{table}

\subsection{Conformer}

Our next experiment replaces the multi-head attention layers in the Transformer model with Conformer layers. Our model contains $14$ Conformer blocks with an embedding dimension of $512$, and $8$ attention heads. The resulting model has the same number of parameters as the Wav2vec 2.0 model. 
Table \ref{tab:lconv} shows that the Conformer block improves performance by $10\%$ relative when fine-tuned on the 100 hour labeled subset.

\subsection{Context and Target MLPs} \label{sec:exp_mlp}
Our final ablation experiments involve adding context and target MLPs to the wav2vec 2.0 model. We experimented with adding each MLP independently, and also the two MLPs together. For the latter case, we also report experiments with different number of MLP layers.
These MLPs contain a batch normalization in each layer, with ReLU activation functions.

The results shown in Table \ref{tab:lconv} indicate that adding either MLP is helpful by itself, with the context MLP being more beneficial than the target MLP. Including both MLPs is more beneficial than adding either MLP by itself. 
Table \ref{tab:lconv} also shows that increasing the number of hidden layers in the MLPs helps. 
Our best model contains $4$ hidden layers in each MLP and obtains a 15\% lower WER when fine-tuned on the 1 hour dataset, and 6\% when fine-tuned on the 100 hour dataset.

%% file: sections/conclusion.tex
\section{Conclusion}
\label{sec:conclusion}

In this work, we showed that data augmentation can be used with Wav2Vec 2.0, a strong self-supervised algorithm for pre-training on speech data. This extends a prior study that focused on the simpler CPC algorithm \cite{kharitonov2021dataug}.
We also presented a number of architectural improvements to the Wav2Vec 2.0 model: we replaced the Transformer blocks with Conformer blocks; we added lightweight and dynamic convolutions to the feature encoder; and finally, we added MLPs on top of the latent speech representations and context vectors to improve the SSL objective.
With all of these changes, our model obtains a combined 13\% lower WER compared to Wav2vec 2.0. 

Prior work has shown that the Wav2Vec 2.0 model can be extended to learn cross-lingual representations from multiple languages \cite{conneau2020unsupervised}. We hypothesize that applying our proposed model for cross-lingual representation learning in a similar manner could significantly benefit languages with limited speech data. We leave this exploration for future work.

%% file: interspeech_main.bbl
\begin{thebibliography}{10}
\providecommand{\url}[1]{#1}
\csname url@samestyle\endcsname
\providecommand{\newblock}{\relax}
\providecommand{\bibinfo}[2]{#2}
\providecommand{\BIBentrySTDinterwordspacing}{\spaceskip=0pt\relax}
\providecommand{\BIBentryALTinterwordstretchfactor}{4}
\providecommand{\BIBentryALTinterwordspacing}{\spaceskip=\fontdimen2\font plus
\BIBentryALTinterwordstretchfactor\fontdimen3\font minus
  \fontdimen4\font\relax}
\providecommand{\BIBforeignlanguage}[2]{{%
\expandafter\ifx\csname l@#1\endcsname\relax
\typeout{** WARNING: IEEEtran.bst: No hyphenation pattern has been}%
\typeout{** loaded for the language `#1'. Using the pattern for}%
\typeout{** the default language instead.}%
\else
\language=\csname l@#1\endcsname
\fi
#2}}
\providecommand{\BIBdecl}{\relax}
\BIBdecl

\bibitem{oord2018cpc}
A.~v.~d. Oord, Y.~Li, and O.~Vinyals, ``Representation learning with
  contrastive predictive coding,'' \emph{arXiv}, vol. abs/1807.03748, 2018.

\bibitem{schneider2019wav2vec}
S.~Schneider, A.~Baevski, R.~Collobert, and M.~Auli, ``wav2vec: Unsupervised
  pre-training for speech recognition,'' in \emph{Proc. Interspeech}, 2019.

\bibitem{harwath2019learning}
D.~Harwath, W.-N. Hsu, and J.~Glass, ``Learning hierarchical discrete
  linguistic units from visually-grounded speech,'' in \emph{Proc. ICLR}, 2020.

\bibitem{chung2019apc}
Y.-A. Chung, W.-N. Hsu, H.~Tang, and J.~R. Glass, ``An unsupervised
  autoregressive model for speech representation learning,'' \emph{arXiv}, vol.
  abs/1904.03240, 2019.

\bibitem{pascual2019learning}
S.~Pascual, M.~Ravanelli, J.~Serrà, A.~Bonafonte, and Y.~Bengio, ``Learning
  problem-agnostic speech representations from multiple self-supervised
  tasks,'' \emph{arXiv}, 2019.

\bibitem{ko2015audio}
T.~Ko, V.~Peddinti, D.~Povey, and S.~Khudanpur, ``Audio augmentation for speech
  recognition,'' in \emph{Sixteenth Annual Conference of the International
  Speech Communication Association}, 2015.

\bibitem{amodei2016deepspeech}
D.~Amodei, S.~Ananthanarayanan, R.~Anubhai, J.~Bai, and et~al., ``Deep speech
  2: End-to-end speech recognition in english and mandarin,'' in \emph{Proc.
  ICML}, 2016.

\bibitem{kharitonov2021dataug}
E.~Kharitonov, M.~Rivière, G.~Synnaeve, L.~Wolf, P.-E. Mazaré, M.~Douze, and
  E.~Dupoux, ``Data augmenting contrastive learning of speech representations
  in the time domain,'' in \emph{IEEE Spoken Language Technology Workshop
  (SLT)}, 2021.

\bibitem{baevski2020wav}
A.~Baevski, H.~Zhou, A.~Mohamed, and M.~Auli, ``wav2vec 2.0: A framework for
  self-supervised learning of speech representations,'' in \emph{Proc.
  NeurIPS}, 2020.

\bibitem{conneau2020unsupervised}
A.~Conneau, A.~Baevski, R.~Collobert, A.~Mohamed, and M.~Auli, ``Unsupervised
  cross-lingual representation learning for speech recognition,'' \emph{arXiv},
  2020.

\bibitem{wu2019pay}
F.~Wu, A.~Fan, A.~Baevski, Y.~N. Dauphin, and M.~Auli, ``Pay less attention
  with lightweight and dynamic convolutions,'' 2019.

\bibitem{gulati2020conformer}
A.~Gulati, J.~Qin, C.-C. Chiu, N.~Parmar, Y.~Zhang, and et~al., ``Conformer:
  Convolution-augmented transformer for speech recognition,'' \emph{arXiv},
  2020.

\bibitem{zhang2020pushing}
Y.~Zhang, J.~Qin, D.~S. Park, W.~Han, C.-C. Chiu, R.~Pang, Q.~V. Le, and Y.~Wu,
  ``Pushing the limits of semi-supervised learning for automatic speech
  recognition,'' \emph{arXiv}, 2020.

\bibitem{he2019momentum}
K.~He, H.~Fan, Y.~Wu, S.~Xie, and R.~Girshick, ``Momentum contrast for
  unsupervised visual representation learning,'' \emph{arXiv}, vol.
  abs/1911.05722, 2019.

\bibitem{chen2020simple}
T.~Chen, S.~Kornblith, M.~Norouzi, and G.~Hinton, ``A simple framework for
  contrastive learning of visual representations,'' \emph{arXiv}, vol.
  abs/2002.05709, 2020.

\bibitem{baevski2019vqwav2vec}
A.~Baevski, S.~Schneider, and M.~Auli, ``vq-wav2vec: Self-supervised learning
  of discrete speech representations,'' in \emph{Proc. ICLR}, 2020.

\bibitem{baevski2019effectiveness}
A.~Baevski, M.~Auli, and A.~Mohamed, ``Effectiveness of self-supervised
  pre-training for speech recognition,'' \emph{arXiv}, vol. abs/1911.03912,
  2019.

\bibitem{vaswani2017transformer}
A.~Vaswani, N.~Shazeer, N.~Parmar, J.~Uszkoreit, L.~Jones, and et~al.,
  ``Attention is all you need,'' in \emph{Proc. NIPS}, 2017.

\bibitem{devlin2018bert}
J.~Devlin, M.-W. Chang, K.~Lee, and K.~Toutanova, ``Bert: Pre-training of deep
  bidirectional transformers for language understanding,'' \emph{arXiv}, vol.
  abs/1810.04805, 2018.

\bibitem{jang2017categorical}
E.~Jang, S.~Gu, and B.~Poole, ``Categorical reparameterization with
  gumbel-softmax,'' 2017.

\bibitem{jegou2011ieee}
H.~Jegou, M.~Douze, and C.~Schmid, ``Product quantization for nearest neighbor
  search,'' \emph{IEEE Trans. on PAMI}, 2011.

\bibitem{jang2016gumbel}
E.~Jang, S.~Gu, and B.~Poole, ``Categorical reparameterization with
  gumbel-softmax,'' \emph{arXiv}, vol. abs/1611.01144, 2016.

\bibitem{howardMobileNets}
\BIBentryALTinterwordspacing
A.~G. Howard, M.~Zhu, B.~Chen, D.~Kalenichenko, W.~Wang, T.~Weyand,
  M.~Andreetto, and H.~Adam, ``Mobilenets: Efficient convolutional neural
  networks for mobile vision applications,'' \emph{CoRR}, vol. abs/1704.04861,
  2017. [Online]. Available: \url{http://arxiv.org/abs/1704.04861}
\BIBentrySTDinterwordspacing

\bibitem{wu2021performanceefficiency}
F.~Wu, K.~Kim, J.~Pan, K.~Han, K.~Q. Weinberger, and Y.~Artzi,
  ``Performance-efficiency trade-offs in unsupervised pre-training for speech
  recognition,'' 2021.

\bibitem{ott2019fairseq}
M.~Ott \emph{et~al.}, ``fairseq: A fast, extensible toolkit for sequence
  modeling,'' in \emph{Proc. NAACL Sys. Demo.}, 2019.

\bibitem{panayotov2015librispeech}
V.~Panayotov, G.~Chen, D.~Povey, and S.~Khudanpur, ``Librispeech: an asr corpus
  based on public domain audio books,'' in \emph{Proc. ICASSP}, 2015.

\bibitem{kahn2020librilight}
J.~Kahn and et~al., ``Libri-light: A benchmark for asr with limited or no
  supervision,'' in \emph{Proc. ICASSP}, 2020.

\bibitem{graves2006connectionist}
A.~Graves, S.~Fern{\'a}ndez, F.~Gomez, and J.~Schmidhuber, ``Connectionist
  temporal classification: labelling unsegmented sequence data with recurrent
  neural networks,'' in \emph{ICML}, 2006.

\bibitem{pratap2019wav2letter}
V.~Pratap, A.~Hannun, Q.~Xu, J.~Cai, J.~Kahn, G.~Synnaeve, V.~Liptchinsky, and
  R.~Collobert, ``Wav2letter++: A fast open-source speech recognition system,''
  in \emph{Proc. ICASSP}, 2019.

\bibitem{snyder2015musan}
D.~Snyder, G.~Chen, and D.~Povey, ``Musan: A music, speech, and noise corpus,''
  2015.

\end{thebibliography}
